\documentclass[sigconf,authorversion,nonacm]{acmart}
\usepackage{multirow}
\usepackage{xcolor}
\usepackage{adjustbox} 
\usepackage{graphicx}
\usepackage{enumitem}
\usepackage{float}       
\usepackage{caption}     
\usepackage{dblfloatfix}   

\AtBeginDocument{%
  }

\setcopyright{acmlicensed}
\copyrightyear{2025}
\acmYear{2025}
\acmDOI{XXXXXXX.XXXXXXX}
\acmConference[ICAIF '25]{The 6th ACM International Conference on AI in Finance}{November 15--18, 2025}{Singapore}
\acmISBN{978-1-4503-XXXX-X/2018/06}

\begin{document}

\title{TradingGroup: A Multi‑Agent Trading System with Self‑Reflection and Data‑Synthesis}


\author{Feng Tian}
\affiliation{%
 \institution{The University of New South Wales}
 \city{Sydney}
 \country{Australia}}
\email{feng.tian1@student.unsw.edu.au}

\author{Flora D. Salim}
\affiliation{%
  \institution{University of New South Wales}
  \city{Sydney}
  \country{Australia}
}
\email{flora.salim@unsw.edu.au}
\orcid{0000-0002-1237-1664}

 \author{Hao Xue}
\affiliation{%
 \institution{The University of New South Wales}
 \city{Sydney}
 \country{Australia}}
 \email{hao.xue1@unsw.edu.au}
 \orcid{0000-0003-1700-9215}

\renewcommand{\shortauthors}{Feng Tian et al.}

\begin{abstract}
Recent advancements in large language models (LLMs) have enabled powerful agent-based applications in finance, particularly for sentiment analysis, financial report comprehension, and stock forecasting. However, existing systems often lack inter-agent coordination, structured self-reflection, and access to high-quality, domain-specific post-training data such as data from trading activities including both market conditions and agent decisions. These data are crucial for agents to understand the market dynamics, improve the quality of decision-making and promote effective coordination. We introduce TradingGroup, a multi-agent trading system designed to address these limitations through a self-reflective architecture and an end-to-end data-synthesis pipeline. TradingGroup consists of specialized agents for news sentiment analysis, financial report interpretation, stock trend forecasting, trading style adaptation, and a trading decision making agent that merges all signals and style preferences to produce buy, sell or hold decisions. Specifically, we design self-reflection mechanisms for the stock forecasting, style, and decision-making agents to distill past successes and failures for similar reasoning in analogous future scenarios and a dynamic risk-management model to offer configurable dynamic stop-loss and take-profit mechanisms. In addition, TradingGroup embeds an automated data-synthesis and annotation pipeline that generates high-quality post-training data for further improving the agent performance through post-training. Our backtesting experiments across five real-world stock datasets demonstrate TradingGroup’s superior performance over rule-based, machine learning, reinforcement learning, and existing LLM-based trading strategies.
\end{abstract}

\ccsdesc[500]{Applied computing~Economics}
\ccsdesc[500]{Applied computing~Forecasting}
\ccsdesc[500]{Computing methodologies~Natural language processing}

\keywords{Large Language Models, Multi-Agent System, Quantitative Trading, Synthetic Data, Parameter-Efficient Fine-Tuning}


\maketitle

\section{Introduction}

Quantitative trading has evolved beyond traditional technical indicators to incorporate multimodal signals such as real-time news, corporate disclosures, and social media sentiment. 
This transformation demands greater robustness and interpretability from trading frameworks which makes it difficult for early rule-based strategies to adapt to the frequently changing market structure. Although reinforcement learning (RL) has the advantage of adaptability, it still struggles to achieve long-term stable profits in real trading scenarios due to limitations such as lack of semantic reasoning, massive training data requirements, and poor generalization in new environments~\cite{zhang2024multimodal}.
With the development of LLMs' reasoning capabilities, financial LLMs such as Fin-R1~\cite{liu2025fin} leverage distillation from DeepSeek-R1~\cite{guo2025deepseek} to obtain high-quality financial reasoning data, and adopt a two-stage training framework combining SFT and RL to achieve strong performance with only 7B parameters. Meanwhile, with the explosion of agentic framework and the enhancement of the tool-use capabilities of the LLMs, excellent quantitative trading agents have emerged such as FinMem~\cite{yu2024finmem} introduces layered memory and role profiles for better interpretability and extensibility; FinAgent~\cite{zhang2024multimodal} unifies multimodal perception, tool use, and self‑reflection; TradingAgents~\cite{xiao2024tradingAgents} demonstrates that a collaborative, game‑theoretic multi‑agent system (MAS) framework is viable for quantitative trading.

However, existing agentic systems still face two key limitations.
First, most rely on static, domain-specific datasets and lack access to real-time, end-to-end trading data, limiting their ability to adapt to dynamic market conditions.
Post-training data describing agent behaviors that captures both market conditions and the corresponding agent decisions is valuable in fine-tuning financial LLMs to systematically improve through experience.
Second, current agentic system has not yet established a self-reflection mechanism that can accurately obtain past transaction experience, deeply integrate multi-agent logs and performance indicators, and directly optimize the workflow of each agent. Current self-reflection methods usually rely on Retrieval-Augmented Generation (RAG~\cite{lewis2020retrieval}) to retrieve relevant cases from historical. However, it fails to integrate execution outcomes, agent logs, and risk signals into a coherent feedback loop. Therefore, in specific situations, it is difficult to accurately distinguish which historical decisions are ``good'' or ``bad'' and hard to make strong adjustments and optimizations to trading strategies based on these experiences.
Hence, this work aims to address two key research questions:
(i) How can a multi-agent trading system integrate performance metrics, agent logs, and risk signals to enable effective self-reflection and strategy optimization?
(ii) How to automatically collect and label trading‑process data to provide high‑quality post-training samples for fine‑tuning base LLMs to further improving the performance?

To address these challenges, we propose TradingGroup, a trading MAS with five collaborative agents and a risk management module. It has a reflection mechanism that enables dynamic style switching, price forecasting and strategy optimization. TradingGroup also embeds an end‑to‑end data‑synthesis pipeline that automatically evaluates and labels trading cases and outputs instruction data suitable for LLM fine‑tuning. 
In summary, our core contributions are:
(1) We propose a novel multi-agent chain trading system with role-specific reasoning capabilities.
(2) We specifically introduce a self-reflection mechanism to provide multi‑level self‑correction across multiple agents and a dynamic risk management module to flexibly adjust stop-loss and take-profit levels as well as position sizing.
(3) We design a data-synthesis pipeline to automatically generate and collect agents' working data for establishing dataset to finetune LLMs. Experiments show that the resulting dataset improves model performance within the backtesting framework.


\section{Related Work}

\subsection{General LLMs and Financial LLMs}
Large language models are moving from text understanding to complex reasoning. Representative reasoning models include OpenAI’s o3/o4-mini~\cite{o3o4systemcard} and DeepSeek-R1, while the open‑source LLM Qwen3 can switch smoothly between thinking mode and non‑thinking mode~\cite{yang2025qwen3}. And with the advances in pre‑training and post‑training, researchers fine‑tune general LLMs on financial domain, equipping them with strong reasoning skills for finance. FinGPT~\cite{yang2023fingptopensourcefinanciallarge} open‑sources a full training and evaluation pipeline for financial LLMs and boosts QA performance with RLHF (Reinforcement Learning from Human Feedback~\cite{ouyang2022traininglanguagemodelsfollow}). Fin‑R1 achieves reasoning comparable to much larger LLMs with only 7B parameters via two‑stage training. FinLlama~\cite{konstantinidis2024finllama} focuses on the task of financial sentiment classification: built on Llama2‑7B, it applies LoRA~\cite{hu2022lora} with 8‑bit quantisation, providing a low‑compute, quick‑to‑deploy paradigm for training financial sentiment LLMs.

However, the limitation of the financial LLMs is that the data they use for training is often constructed based on fixed domain knowledge. They lack real‑time trading experience that contains the current state of the market. This limits the model’s capacity to reason through real-time situations and hinders optimisation during live trading. Our TradingGroup fills this gap with an agent data‑synthesis and auto‑labelling pipeline.

\subsection{General Agents and Financial Agents}
Recently, the research community start to focus on developing and improving the agentic capabilities of LLMs. MetaGPT~\cite{hong2023metagpt} enables LLMs to call external tools, break down tasks and collaborate to achieve goals. OpenAI’s Deep Research~\cite{deepresearch2025} was trained on browsing data and refined via reinforcement learning to search and analyze web content. The research of agent for trading is emerging as well: FinMem adds layered memory and role profiles in the agentic framework, making trading decisions more interpretable; FinAgent integrates multimodal LLMs, tool retrieval and self-reflection. TradingAgents adopts a game‑theoretic MAS that splits roles such as analyst, trader, and risk controller. Building on the proven strengths of reflection modules (FinAgent) and MAS architectures (TradingAgents), TradingGroup integrates self-reflection into its trading-decision, price‑forecasting, and style-preference agents, and couples them with a dynamic risk‑management module, achieving higher stability and backtesting performance.


\begin{figure*}[htbp]
  \centering
  \includegraphics[width=.88\textwidth]{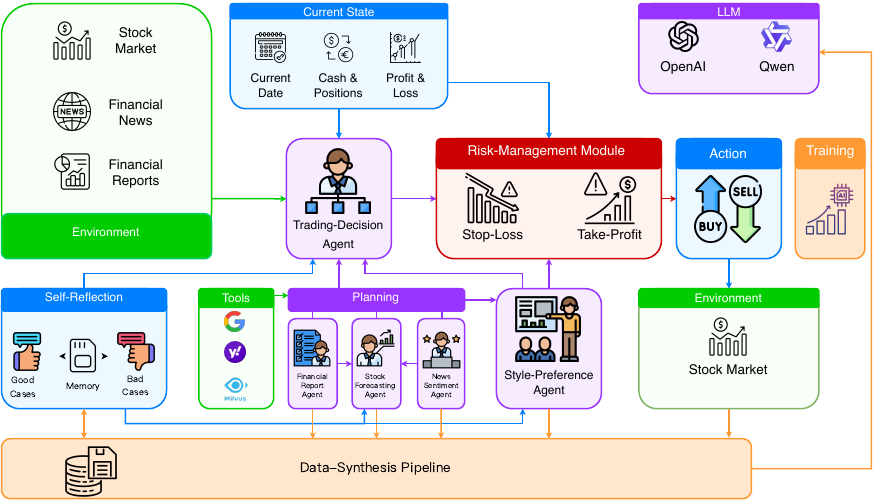}
  \caption{The overall framework architecture of the proposed multi-agent TradingGroup.}
  \label{fig:tradinggroup-architecture}
\end{figure*}

\section{TradingGroup Framework}
As shown in Figure 1, TradingGroup is composed of 4 functional parts working together: five types of agents, dynamic risk management module, self-reflection mechanism and data-synthesis pipeline. Specifically, the \textit{News-Sentiment Agent} is responsible for filtering financial News and summarizing it into the overall market sentiment score, aiming to help the system better capture changes in market sentiment. \textit{Financial-Report Agent} uses the optimized RAG to analyze the quarterly or annual reports of enterprises and provides current financial states for other agents. \textit{Stock-Forecasting Agent} predicts stock price trend by combining the self-reflection mechanism and the technical factors in the stock price with the analysis results of other agents. \textit{Style-Preference Agent}, based on the current status of the account, historical trading performance and other agent's analysis, outputs the current preferred trading style and confidence scores after self-reflection. \textit{Trading-Decision Agent} integrates all upstream analyses, reviews past profit and loss cases, and after self-reflection, provides the final trading actions. Risk-Management Module is designed to dynamically adjust the take-profit and stop-loss thresholds and manage risks in real time. The Self-Reflection Mechanism is integrated into the Stock-Forecasting, Style-Preference, and Trading-Decision Agents, allowing them to leverage past experiences to enhance decision-making in similar market conditions. Data-Synthesis Pipeline automatically generates the working data and reward parameters of the agents, providing data support for post-training. The output of each agent is accompanied by an explanation. When using DeepSeek-R1 or Qwen3 reasoning, the Chain-of-Thought (CoT~\cite{wei2022chain}) trajectory is saved to provide high-quality data for data distillation~\cite{guo2025deepseek}.

\subsection{News-Sentiment Agent}


Due to the complexity and noise of news, the accurate judgment of the real market sentiment is challenging. To address this,  we designe a financial news workflow that includes retrieval, re-ranking, filtering and scoring modules, aiming to help agents understand the current market environment more accurately and efficiently. To efficiently obtain real-time news of the current market, we designe an MCP (Model Context Protocol~\cite{mcp2025}) client that enables our News-Sentiment Agent using online search tools such as Serper MCP Server~\cite{serpermcp2025}. To get the most important news, we compute an influence score by Qwen3-Reranker-0.6B~\cite{zhang2025qwen3} for every news:
$N_t = 0.55\,\mathrm{base}(t) + 0.25\,\mathrm{prob}(t) + 0.20$,
where the final importance score \(N_t\) for the \(t\)-th news item is a weighted sum of three components.  Here, \(\mathrm{base}(t)\) denotes the importance score computed by keyword and length‐based rules, \(\mathrm{prob}(t)\) is the probability output by the Qwen3‐Reranker that the news is impactful, and the constant \(0.20\) serves as a minimum score bias.

Based on the influence score, we select the top-k news and perform vector similarity deduplication using Qwen3-Embedding-0.6B~\cite{zhang2025qwen3}. Then, the LLM analyzes each news in parallel and aggregates overall market sentiment score and analysis. By using these methods, we effectively filter out the noise, enabling the News-Sentiment Agent to analyze current market more accurately and efficiently.


\subsection{Financial-Report Agent}
The annual and quarterly reports of enterprises contain key indicators such as revenue, performance guidance, and risk warnings. These financial data can help agents better understand the company's operating conditions, potential risks and opportunities, thereby providing background support for decision-making. To accurately obtain the relevant data in the company's financial reports that may have an impact on the market and conduct analysis, we have designed an improved RAG method with Milvus~\cite{2021milvus}.

\noindent\textbf{Hybrid Retrieval.} In order to filter the content and obtain information related to financial indicators, after sliding‑window sentence chunking, we perform dense retrieval with Qwen3‑Embedding‑0.6B to capture semantically similar chunks, while BGE‑M3‑0.6B~\cite{chen2024bge} is used to improve keyword matches:
\begin{equation}
  \label{eq:hybrid_score}
  H_i = w_d \times \mathrm{dense}(q,i)
        + w_s \times \mathrm{sparse}(q,i)
\end{equation}
where \(w_d=1.0\) and \(w_s=0.8\). Here, \(\mathrm{dense}(q,i)\) denotes the cosine similarity between the query embedding and the \(i\)-th document embedding, both produced by Qwen3-Embedding-0.6B; and \(\mathrm{sparse}(q,i)\) denotes the inner product between the query embedding and the \(i\)-th document embedding, both produced by BGE-M3-0.6B. We select the top 10 chunks with the highest scores as the results of hybrid retrieval.

\noindent\textbf{LLM Re-ranking.} To identify technical indicators most relevant to stock price, we use Qwen3-Reranker-0.6B to re-rank passages from hybrid retrieval. A yes/no prompt asks whether each passage contains price-related factors, and the model's final-token logits serve as re-ranking signals. The top six passages are used as references for the core LLM to summarize relevant technical indicators.

\subsection{Stock-Forecasting Agent}
Historical price is rich in technical signals and serves as the foundation for quantitative decision-making. The Stock-Forecasting Agent in TradingGroup not only reads market data to calculate technical indicators, but also integrates the outputs of other agents to deliver a more robust trend assessment from the three dimensions of price, sentiment and fundamentals. To fetch the real-time market data, we use yfinance library~\cite{yfinance} in the online mode. To help the agent better analyze the current market, we have combined the agent with the financial technology indicators. In our Stock-Forecasting Agent, an RSI-14 (Relative Strength Index~\cite{wilder1978new}) above 70 signals overbought and below 40 signals oversold. Distance to 20-day High/Low(\%) is defined as the percentage distance between the current price and the past 20-day highest (or lowest) close. It is used to identify potential breakthroughs or bottom breaks. Distance to 20-day SMA(\%) is defined as the percentage deviation of the current close price from the 20-day simple moving average. A 0 to -3\% deviation with an RSI less than 40 is considered a healthy pullback. 20-day High/Low Flag is a boolean indicator which is set to true when the current closing price establishes a new 20-day closing high or low. HV-10(\%) (Historical Volatility of 10 days) is defined as the annualized volatility of the logarithmic return rate over the last 10 days. If it is less than 20\%, it may be in a sideways movement. It is used for dynamic threshold calculation. 

Simplified ATR-20(\%) (Average True Range of 20 days~\cite{wilder1978new}) is calculated using the standard deviation of 20-day close-to-close log returns (we use the close price here to adapt to different datasets). This metric serves as an approximation of the fluctuation range: 
\begin{equation}
  \mathrm{SimpliedATR}_{20}(\%) 
  = 100\;\sqrt{\frac{1}{20}\sum_{t=T-19}^{T}\Bigl(\ln\!\frac{P_t}{P_{t-1}} - \overline{r}_{20}\Bigr)^{2}}
\end{equation}
Here, \(P_t\) remains the closing price on day \(t\), and \(\overline{r}_{20}\) is the arithmetic mean of the last twenty log‐returns. Taking the square root of the average squared deviation gets the standard deviation of these returns.


\smallskip
\noindent\textbf{Self-Reflection Mechanism:} Inspired by the ReAct~\cite{yao2023react} framework and FinAgent, but the different thing is that our self-reflection mechanism is build on the TradingGroup data pipeline. In the Stock-Forecasting Agent, the self-reflection mechanism extracts recent successful and failed prediction cases through the data pipeline, summarises their patterns and root causes, injects the conclusions into the LLM context, and prompts the model to self-correct in similar market conditions (see Figure 2). Our self-reflection mechanism can more accurately obtain historical successful and failed cases and conduct efficient analysis.

\smallskip
\noindent\textbf{Hybrid Gate:} If the RSI is overheated and the distance to a valid breakout still exceeds the breakout threshold, the agent forces a sideways classification to avoid chasing tops, which is called hard interception. 
The breakout threshold (\%) is inspired by the ATR Channel Breakout strategy~\cite{liu2015empirical}, with a minimum of 1\%. This threshold distinguishes a mere approach to a recent high from a valid breakout: $\mathrm{BreakoutThreshold}(\%) 
  = \max\!\bigl(1\%,\;0.5 \times \mathrm{SimpliedATR}_{20}(\%)\bigr)$.
For all other situations we move to a soft pass. If the LLM's uptrend probability exceeds a fixed threshold and at least one of two bullish technical patterns is triggered, we declare an uptrend (soft pass). Otherwise, if the downtrend probability is above 0.55, we assign a downtrend. All remaining cases default to sideways. This hybrid gating mechanism preserves the LLM's sensitivity to news and fundamentals while constraining predictions with objective rules, preventing overfitting and extreme misclassifications.


\subsection{Style-Preference Agent}
In the FinArena~\cite{xu2025finarena}, investor preferences are classified into four categories and user can set different preference, while in the FinMem, style prompts are integrated into the context and automatically switched with different stocks. In TradingGroup, we integrate style preference, dynamic risk management, and self-reflection, enhancing the influence of trading styles on the entire system. The Style-Preference Agent combines other agents' analysis and uses self-reflection to analyze the recent performance with current state to choose the most suitable trading style.

The self-reflection mechanism plays a key role in the Style-Preference Agent by analyzing historical trading records, including the styles adopted and corresponding profit and loss. By comparing these with the agent's current capital, holdings, and PnL status, it dynamically adjusts the preferred trading style (see Figure 2). This reflective process helps the agent identify which style has been more effective under similar conditions, enabling it to make more informed and adaptive decisions in real time.


This style classification directly affects the execution intensity of trading actions: when buying, aggressive and balanced styles invest all available cash, while the conservative style uses only 50\%. When selling, the aggressive style halves the position, whereas balanced and conservative styles fully liquidate it.

Additionally, the Style-Preference Agent writes different historical volatility coefficients to the dynamic risk management module, thereby adjusting the take-profit and stop-loss thresholds.

\subsection{Trading-Decision Agent}
The Trading-Decision Agent is at the top of the TradingGroup system. It is responsible for integrating the information from the Stock-Forecasting, News-Sentiment, Financial-Report and Style-Preference Agent with current account status, volatility constraints and historical performance. It then outputs the Buy/Hold/Sell action of the day with explanations, ensuring that the final decisions it makes are based on the comprehensive information of the current market and the entire TradingGroup system.

For self-reflection, leveraging the data-synthesis pipeline, we label each decision from the past 20 trading days with its actual market outcome, select both successful and failed cases, and automatically compile an "experience summary." This text is prepended to the final prompt, prompting the LLM to explicitly review and correct past mistakes before producing a new action (see Figure 2).

\subsection{Risk-Management Module}
Risk management is the fundamental guarantee to keep a quantitative system alive for long-term. Unlike TradingAgents, which uses risk management team, our TradingGroup adopts a scheme that pairs style-tiered dynamic thresholds with hard intercept. Firstly, the module calculates the simplified 10-day historical volatility. Then, different coefficients are injected according to the trading style to obtain adaptive take-profit ($T_{\mathrm{TP}}$)/stop-loss ($T_{\mathrm{SL}}$) threshold:
\begin{equation}
  T_{\mathrm{SL}} = m_{s}^{\mathrm{sl}}\;\sigma_{d,10}, 
  \quad
  T_{\mathrm{TP}} = m_{s}^{\mathrm{tp}}\;\sigma_{d,10}
\end{equation}
Here, \(\sigma_{d,10}\) is the unannualized standard deviation of daily log-returns over the past 10 trading days. The style-dependent multipliers \(m_{s}^{\mathrm{sl}}\) and \(m_{s}^{\mathrm{tp}}\) come from the configuration for style \(s\). These give the adaptive stop-loss threshold \(T_{\mathrm{SL}}\) and take-profit threshold \(T_{\mathrm{TP}}\), respectively. Continuous monitoring of each position’s unrealized PnL (Profit and Loss) percentage then triggers a forced sell when PnL is less than or equal to $-T_{\mathrm{SL}}$ or a profit-take when PnL is greater than or equal to $T_{\mathrm{TP}}$.

We adopt this hard-intercept mechanism that monitors PnL and immediately triggers a stop-lossing or profit-taking action once the absolute value exceeds threshold. This decisive control strategy enables timely responses to risks, with execution size dynamically adjusted by trading style.

\subsection{Data-Synthesis Pipeline}

\begin{figure*}[!t]
  \centering
  \includegraphics[width=.965\textwidth]{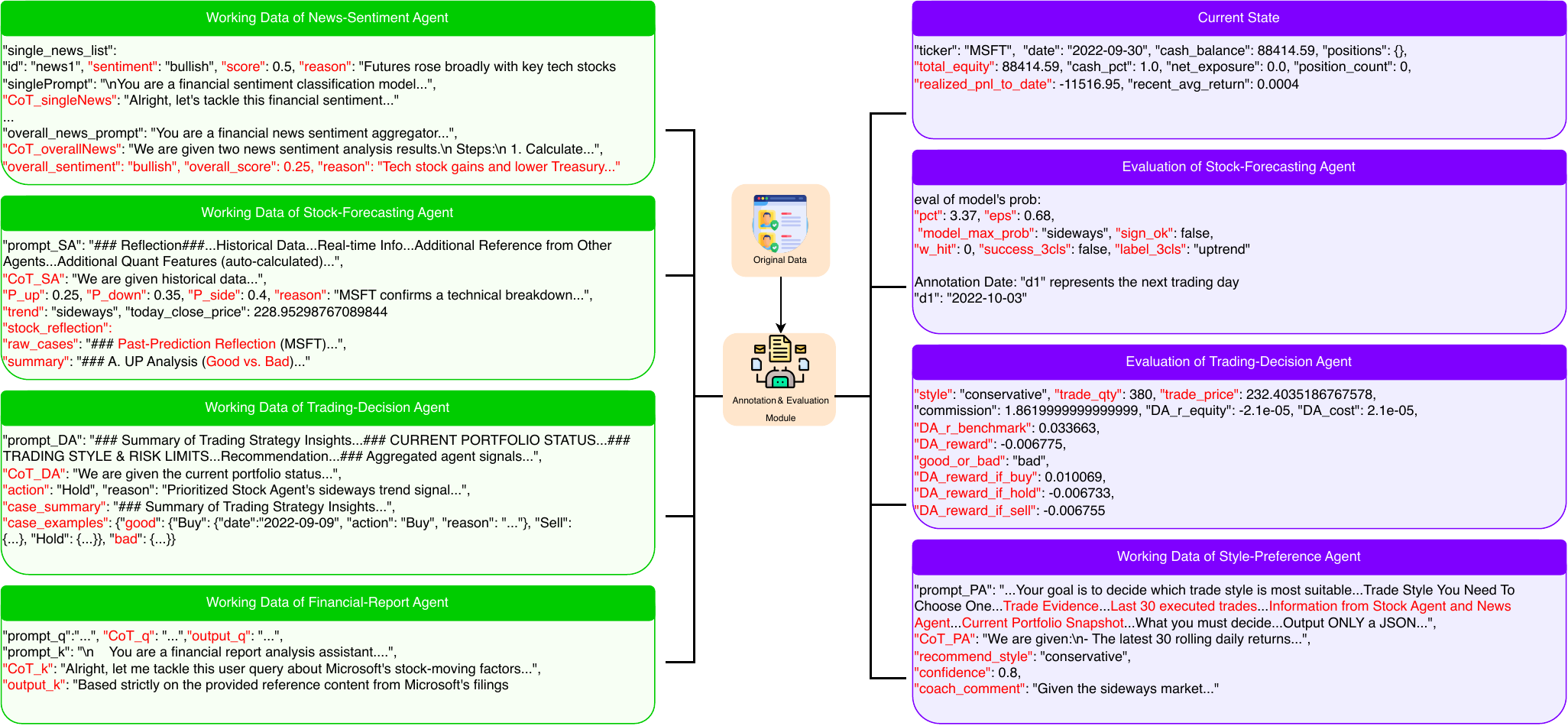}
  \caption{Working data flow in the data–synthesis and annotation \& evaluation pipeline of TradingGroup.}
  \label{fig:data-pipeline}
\end{figure*}

To support agent fine-tuning and self-reflection, we introduce an end-to-end data-synthesis pipeline within TradingGroup (see Figure 1). Equipped with automatic labeling, the pipeline instantly converts the raw records into data samples that can be used for SFT or RL.
Specifically, the pipeline collects the following for each agent: the input and output texts; the account state for the day (date, positions, available cash, trading style, etc.); and the full CoT generated during LLM inference (if applicable) (see Figure 2).

Additionally, we design dedicated labels and reward parameters for two key agents. These reward parameters are not only useful for self-reflection, but are also critical for data selection. Specifically, we use them to filter high-quality samples for SFT: for the Trading-Decision Agent, only actions with positive \( \mathrm{reward}_a \) are kept; for the Stock-Forecasting Agent, predictions with high \( w_{\mathrm{hit}} \) are selected based on directional accuracy, confidence, and signal clarity. This reward-guided filtering ensures that only meaningful, well-grounded samples are used for SFT.

For Stock-Forecasting Agent, we automatically assign a correct or incorrect label to the prediction made for the preceding trading day. Using the actual price movement, we then compute the necessary reward parameters for self-reflection and future training. The formula of reward parameters are as follows: $\epsilon = \max\!\Bigl(
    \alpha \times \frac{1}{20}\sum_{t=T-19}^{T}\bigl|\ln\tfrac{P_t}{P_{t-1}}\bigr|,
    \;\epsilon_{\min}
  \Bigr)$
where \(P_t\) is the closing price on day \(t\), \(\ln(P_t/P_{t-1})\) is the daily log-return. The average
measures the past 20 trading days’ daily volatility; \(\alpha\) scales that estimate; and \(\epsilon_{\min}\) is the minimum value. The resulting \(\epsilon\) is the threshold: any return whose absolute value is less or equal to \(\epsilon\) will be classified as sideways. 
The realized one-day return is defined as 
$\mathrm{pct} = \frac{P_{d_1}}{P_{d_0}} - 1,$
where \(P_{d_0}\) and \(P_{d_1}\) are the closing prices on the prediction date and the next trading day, respectively.

\begin{equation}
w_{\mathrm{hit}} = \mathrm{sign\_ok}\;\times\;\tanh\!\Bigl(\frac{|\mathrm{pct}|}{\epsilon}\Bigr)\;\times\;p_{\mathrm{true}}
\end{equation}
Here, \(\tanh\bigl(|\mathrm{pct}|/\epsilon\bigr)\) scales the return relative to the band into \((-1,1)\), and \(p_{\mathrm{true}}\) is the model's predicted probability for the true label. Here, \(\mathrm{sign\_ok}\) is a binary indicator that equals 1 if and only if the predicted direction is correct (specifically, if the prediction of the max probability is sideways then \(|\mathrm{pct}|\le\epsilon\)). \(w_{\mathrm{hit}}\) represents the weighted hit bonus, rewarding correct directional predictions in proportion to both the strength of the return and the model’s confidence.

For Trading-Decision Agent, based on position changes and realised profit and loss, each Buy, Hold or Sell action is labeled as profitable or loss-making. Using metrics such as account return, the agent then computes the reward parameters: 
\begin{equation}
\mathrm{reward}_a
= r_{\mathrm{eq},a}
  - \beta\,r_{\mathrm{bm}}
   - \gamma\,c_a
 \end{equation}
Here, \(r_{\mathrm{eq},a} = (E_a - E_{\mathrm{prev}})/E_{\mathrm{prev}}\) is the simulated return under action \(a \in \{\text{buy, hold, sell}\}\), where \(E_{\mathrm{prev}}\) is the portfolio equity before the action, and \(E_a\) is the equity after applying the action (including trading and commission). The benchmark return \(r_{\mathrm{bm}} = (P_{\mathrm{curr}} - P_{\mathrm{prev}})/P_{\mathrm{prev}}\) reflects Buy-and-Hold strategy. The transaction cost ratio \(c_a = \mathrm{commission}_a / E_{\mathrm{prev}}\) is penalized separately. \(\beta\) and \(\gamma\) are penalty coefficients (default: \(\beta = 0.2\), \(\gamma = 1.0\)).

For the Trading-Decision Agent, \( \mathrm{reward}_a \) quantifies the effectiveness of Buy, Hold, or Sell actions. For the Stock-Forecasting Agent, \( w_{\mathrm{hit}} \) rewards directionally correct predictions based on actual price movement, volatility-adjusted thresholds, and model confidence. These signals are used to filter high-quality examples for SFT.

\begin{table}[htbp]
\centering
\caption{Quantitative Backtesting (on FINSABER Framework) of TradingGroup and Baseline Strategies, `–' Indicates No Trade Activities; Red (Bold), Blue, and Orange Show the Best, Second-Best, and Third-Best Results}
\resizebox{0.95\linewidth}{!}{ 
\begin{tabular}{cclllll}
\hline
\multicolumn{1}{l|}{Symbol}                  & \multicolumn{1}{l|}{Type}                                                                   & \multicolumn{1}{l|}{Strategy}                                  & SPR↑                         & CR(\%) ↑                      & MDD ↑                          & AV(\%) ↓                      \\ \hline
\multicolumn{1}{c|}{}                        & \multicolumn{1}{c|}{}                                                                       & \multicolumn{1}{l|}{Buy and Hold}                              & -0.342                       & -20.481                       & -52.727                        & 55.908                        \\
\multicolumn{1}{c|}{}                        & \multicolumn{1}{c|}{}                                                                       & \multicolumn{1}{l|}{SMA Cross}                                 & {\color[HTML]{F2BA02} 1.237} & {\color[HTML]{00B0F0} 30.683} & -22.399                        & {\color[HTML]{F2BA02} 33.389} \\
\multicolumn{1}{c|}{}                        & \multicolumn{1}{c|}{}                                                                       & \multicolumn{1}{l|}{WMA Cross}                                 & {\color[HTML]{00B0F0} 1.337} & \textbf{{\color[HTML]{FF0000} 32.924}} & {\color[HTML]{00B0F0} -13.476} & {\color[HTML]{00B0F0} 32.559} \\
\multicolumn{1}{c|}{}                        & \multicolumn{1}{c|}{}                                                                       & \multicolumn{1}{l|}{ATR Band}                                  & -0.595                       & -19.142                       & -39.599                        & 42.161                        \\
\multicolumn{1}{c|}{}                        & \multicolumn{1}{c|}{}                                                                       & \multicolumn{1}{l|}{Bollinger Bands}                           & -0.769                       & -24.749                       & -44.655                        & 45.366                        \\
\multicolumn{1}{c|}{}                        & \multicolumn{1}{c|}{\multirow{-6}{*}{\begin{tabular}[c]{@{}c@{}}Rule\\ Based\end{tabular}}} & \multicolumn{1}{l|}{Turn of The Month}                         & 0.219                        & 3.636                         & \textbf{{\color[HTML]{FF0000} -11.643}} & \textbf{{\color[HTML]{FF0000} 31.043}} \\ \cline{2-7} 
\multicolumn{1}{c|}{}                        & \multicolumn{1}{c|}{}                                                                       & \multicolumn{1}{l|}{ARIMA}                                     & 0.464                        & 10.301                        & -23.524                        & 41.496                        \\
\multicolumn{1}{c|}{}                        & \multicolumn{1}{c|}{\multirow{-2}{*}{Predictor}}                                            & \multicolumn{1}{l|}{XGBoost}                                   & -0.262                       & -14.133                       & -43.457                        & 47.944                        \\ \cline{2-7} 
\multicolumn{1}{c|}{}                        & \multicolumn{1}{c|}{}                                                                       & \multicolumn{1}{l|}{A2C}                                       & -0.343                       & -20.423                       & -52.592                        & 55.859                        \\
\multicolumn{1}{c|}{}                        & \multicolumn{1}{c|}{}                                                                       & \multicolumn{1}{l|}{DDPG}                                      & —                            & —                             & —                              & —                             \\
\multicolumn{1}{c|}{}                        & \multicolumn{1}{c|}{}                                                                       & \multicolumn{1}{l|}{PPO}                                       & —                            & —                             & —                              & —                             \\
\multicolumn{1}{c|}{}                        & \multicolumn{1}{c|}{}                                                                       & \multicolumn{1}{l|}{SAC}                                       & -0.343                       & -20.423                       & -52.592                        & 55.859                        \\
\multicolumn{1}{c|}{}                        & \multicolumn{1}{c|}{\multirow{-5}{*}{FinRL}}                                                & \multicolumn{1}{l|}{TD3}                                       & —                            & —                             & —                              & —                             \\ \cline{2-7} 
\multicolumn{1}{c|}{}                        & \multicolumn{1}{c|}{}                                                                       & \multicolumn{1}{l|}{FinMem (GPT-4o-mini)}                      & 0.312                        & 3.472                         & -30.142                        & 46.32                         \\
\multicolumn{1}{c|}{}                        & \multicolumn{1}{c|}{}                                                                       & \multicolumn{1}{l|}{FinAgent (GPT-4o-mini)}                    & —                            & —                             & —                              & —                             \\
\multicolumn{1}{c|}{\multirow{-16}{*}{TSLA}} & \multicolumn{1}{c|}{\multirow{-3}{*}{LLM}}                                                  & \multicolumn{1}{l|}{\textbf{(Ours)TradingGroup (GPT-4o-mini)}} & \textbf{{\color[HTML]{FF0000} 1.398}} & {\color[HTML]{F2BA02} 25.662} & {\color[HTML]{F2BA02} -15.305} & 34.78                         \\ \hline
\multicolumn{1}{c|}{}                        & \multicolumn{1}{c|}{}                                                                       & \multicolumn{1}{l|}{Buy and Hold}                              & 1.326                        & {\color[HTML]{00B0F0} 43.079} & -20.184                        & 41.523                        \\
\multicolumn{1}{c|}{}                        & \multicolumn{1}{c|}{}                                                                       & \multicolumn{1}{l|}{SMA Cross}                                 & -1.124                       & -17.576                       & -27.29                         & 25.309                        \\
\multicolumn{1}{c|}{}                        & \multicolumn{1}{c|}{}                                                                       & \multicolumn{1}{l|}{WMA Cross}                                 & -0.189                       & -2.289                        & -20.939                        & 22.169                        \\
\multicolumn{1}{c|}{}                        & \multicolumn{1}{c|}{}                                                                       & \multicolumn{1}{l|}{ATR Band}                                  & 0.15                         & 2.992                         & {\color[HTML]{F2BA02} -12.231} & 19.315                        \\
\multicolumn{1}{c|}{}                        & \multicolumn{1}{c|}{}                                                                       & \multicolumn{1}{l|}{Bollinger Bands}                           & -0.558                       & -4.996                        & -13.244                        & {\color[HTML]{00B0F0} 16.754} \\
\multicolumn{1}{c|}{}                        & \multicolumn{1}{c|}{\multirow{-6}{*}{\begin{tabular}[c]{@{}c@{}}Rule\\ Based\end{tabular}}} & \multicolumn{1}{l|}{Turn of The Month}                         & 0.559                        & 8.383                         & {\color[HTML]{00B0F0} -10.641} & {\color[HTML]{F2BA02} 17.194} \\ \cline{2-7} 
\multicolumn{1}{c|}{}                        & \multicolumn{1}{c|}{}                                                                       & \multicolumn{1}{l|}{ARIMA}                                     & \textbf{{\color[HTML]{FF0000} 1.946}} & \textbf{{\color[HTML]{FF0000} 53.64}}  & -13.247                        & 31.591                        \\
\multicolumn{1}{c|}{}                        & \multicolumn{1}{c|}{\multirow{-2}{*}{Predictor}}                                            & \multicolumn{1}{l|}{XGBoost}                                   & 0.734                        & 9.452                         & \textbf{{\color[HTML]{FF0000} -7.197}}  & \textbf{{\color[HTML]{FF0000} 14.386}} \\ \cline{2-7} 
\multicolumn{1}{c|}{}                        & \multicolumn{1}{c|}{}                                                                       & \multicolumn{1}{l|}{A2C}                                       & —                            & —                             & —                              & —                             \\
\multicolumn{1}{c|}{}                        & \multicolumn{1}{c|}{}                                                                       & \multicolumn{1}{l|}{DDPG}                                      & —                            & —                             & —                              & —                             \\
\multicolumn{1}{c|}{}                        & \multicolumn{1}{c|}{}                                                                       & \multicolumn{1}{l|}{PPO}                                       & —                            & —                             & —                              & —                             \\
\multicolumn{1}{c|}{}                        & \multicolumn{1}{c|}{}                                                                       & \multicolumn{1}{l|}{SAC}                                       & 1.325                        & {\color[HTML]{F2BA02} 42.872} & -20.121                        & 41.448                        \\
\multicolumn{1}{c|}{}                        & \multicolumn{1}{c|}{\multirow{-5}{*}{FinRL}}                                                & \multicolumn{1}{l|}{TD3}                                       & —                            & —                             & —                              & —                             \\ \cline{2-7} 
\multicolumn{1}{c|}{}                        & \multicolumn{1}{c|}{}                                                                       & \multicolumn{1}{l|}{FinMem (GPT-4o-mini)}                      & {\color[HTML]{00B0F0} 1.704} & 32.549                        & -13.018                        & 34.766                        \\
\multicolumn{1}{c|}{}                        & \multicolumn{1}{c|}{}                                                                       & \multicolumn{1}{l|}{FinAgent (GPT-4o-mini)}                    & {\color[HTML]{F2BA02} 1.543} & 41.167                        & -20.417                        & 51.03                         \\
\multicolumn{1}{c|}{\multirow{-16}{*}{NFLX}} & \multicolumn{1}{c|}{\multirow{-3}{*}{LLM}}                                                  & \multicolumn{1}{l|}{\textbf{(Ours)TradingGroup (GPT-4o-mini)}} & 1.282                        & 20.458                        & -14.958                        & 30.131                        \\ \hline
\multicolumn{1}{c|}{}                        & \multicolumn{1}{c|}{}                                                                       & \multicolumn{1}{l|}{Buy and Hold}                              & -0.46                        & -13.25                        & -31.546                        & 35.624                        \\
\multicolumn{1}{c|}{}                        & \multicolumn{1}{c|}{}                                                                       & \multicolumn{1}{l|}{SMA Cross}                                 & {\color[HTML]{00B0F0} 0.77}  & {\color[HTML]{00B0F0} 13.265} & -11.653                        & 22.874                        \\
\multicolumn{1}{c|}{}                        & \multicolumn{1}{c|}{}                                                                       & \multicolumn{1}{l|}{WMA Cross}                                 & 0.235                        & 4.194                         & -14.613                        & 22.888                        \\
\multicolumn{1}{c|}{}                        & \multicolumn{1}{c|}{}                                                                       & \multicolumn{1}{l|}{ATR Band}                                  & {\color[HTML]{F2BA02} 0.623} & {\color[HTML]{F2BA02} 11.012} & -15.841                        & 23.273                        \\
\multicolumn{1}{c|}{}                        & \multicolumn{1}{c|}{}                                                                       & \multicolumn{1}{l|}{Bollinger Bands}                           & -0.402                       & -7.1                          & -20.614                        & 26.559                        \\
\multicolumn{1}{c|}{}                        & \multicolumn{1}{c|}{\multirow{-6}{*}{\begin{tabular}[c]{@{}c@{}}Rule\\ Based\end{tabular}}} & \multicolumn{1}{l|}{Turn of The Month}                         & -0.037                       & 0.044                         & -14.892                        & 20.723                        \\ \cline{2-7} 
\multicolumn{1}{c|}{}                        & \multicolumn{1}{c|}{}                                                                       & \multicolumn{1}{l|}{ARIMA}                                     & 0.029                        & 0.553                         & -10.854                        & 24.267                        \\
\multicolumn{1}{c|}{}                        & \multicolumn{1}{c|}{\multirow{-2}{*}{Predictor}}                                            & \multicolumn{1}{l|}{XGBoost}                                   & -0.361                       & -5.295                        & -19.91                         & 22.413                        \\ \cline{2-7} 
\multicolumn{1}{c|}{}                        & \multicolumn{1}{c|}{}                                                                       & \multicolumn{1}{l|}{A2C}                                       & —                            & —                             & —                              & —                             \\
\multicolumn{1}{c|}{}                        & \multicolumn{1}{c|}{}                                                                       & \multicolumn{1}{l|}{DDPG}                                      & —                            & —                             & —                              & —                             \\
\multicolumn{1}{c|}{}                        & \multicolumn{1}{c|}{}                                                                       & \multicolumn{1}{l|}{PPO}                                       & —                            & —                             & —                              & —                             \\
\multicolumn{1}{c|}{}                        & \multicolumn{1}{c|}{}                                                                       & \multicolumn{1}{l|}{SAC}                                       & —                            & —                             & —                              & —                             \\
\multicolumn{1}{c|}{}                        & \multicolumn{1}{c|}{\multirow{-5}{*}{FinRL}}                                                & \multicolumn{1}{l|}{TD3}                                       & —                            & —                             & —                              & —                             \\ \cline{2-7} 
\multicolumn{1}{c|}{}                        & \multicolumn{1}{c|}{}                                                                       & \multicolumn{1}{l|}{FinMem (GPT-4o-mini)}                      & -0.888                       & -4.83                         & {\color[HTML]{F2BA02} -10.281} & {\color[HTML]{00B0F0} 13.505} \\
\multicolumn{1}{c|}{}                        & \multicolumn{1}{c|}{}                                                                       & \multicolumn{1}{l|}{FinAgent (GPT-4o-mini)}                    & 0.52                         & 4.454                         & {\color[HTML]{00B0F0} -5.708}  & \textbf{{\color[HTML]{FF0000} 12.432}} \\
\multicolumn{1}{c|}{\multirow{-16}{*}{AMZN}} & \multicolumn{1}{c|}{\multirow{-3}{*}{LLM}}                                                  & \multicolumn{1}{l|}{\textbf{(Ours)TradingGroup (GPT-4o-mini)}} & \textbf{{\color[HTML]{FF0000} 3.859}} & \textbf{{\color[HTML]{FF0000} 40.458}} & \textbf{{\color[HTML]{FF0000} -2.118}}  & {\color[HTML]{F2BA02} 17.228} \\ \hline
\multicolumn{1}{c|}{}                        & \multicolumn{1}{c|}{}                                                                       & \multicolumn{1}{l|}{Buy and Hold}                              & 0.974                        & {\color[HTML]{00B0F0} 21.128} & -14.165                        & 28.271                        \\
\multicolumn{1}{c|}{}                        & \multicolumn{1}{c|}{}                                                                       & \multicolumn{1}{l|}{SMA Cross}                                 & -0.145                       & -0.291                        & -13.809                        & {\color[HTML]{F2BA02} 16.124} \\
\multicolumn{1}{c|}{}                        & \multicolumn{1}{c|}{}                                                                       & \multicolumn{1}{l|}{WMA Cross}                                 & -0.427                       & -3.421                        & -14.748                        & 16.376                        \\
\multicolumn{1}{c|}{}                        & \multicolumn{1}{c|}{}                                                                       & \multicolumn{1}{l|}{ATR Band}                                  & 1.036                        & 12.981                        & {\color[HTML]{00B0F0} -7.71}   & \textbf{{\color[HTML]{FF0000} 15.007}} \\
\multicolumn{1}{c|}{}                        & \multicolumn{1}{c|}{}                                                                       & \multicolumn{1}{l|}{Bollinger Bands}                           & \textbf{{\color[HTML]{FF0000} 2.115}} & \textbf{{\color[HTML]{FF0000} 31.62}}  & \textbf{{\color[HTML]{FF0000} -3.475}}  & 18.245                        \\
\multicolumn{1}{c|}{}                        & \multicolumn{1}{c|}{\multirow{-6}{*}{\begin{tabular}[c]{@{}c@{}}Rule\\ Based\end{tabular}}} & \multicolumn{1}{l|}{Turn of The Month}                         & -0.036                       & 0.951                         & -11.95                         & {\color[HTML]{00B0F0} 15.09}  \\ \cline{2-7} 
\multicolumn{1}{c|}{}                        & \multicolumn{1}{c|}{}                                                                       & \multicolumn{1}{l|}{ARIMA}                                     & {\color[HTML]{F2BA02} 1.095} & 20.228                        & -12.073                        & 22.993                        \\
\multicolumn{1}{c|}{}                        & \multicolumn{1}{c|}{\multirow{-2}{*}{Predictor}}                                            & \multicolumn{1}{l|}{XGBoost}                                   & 0.989                        & 15.874                        & -11.028                        & 19.503                        \\ \cline{2-7} 
\multicolumn{1}{c|}{}                        & \multicolumn{1}{c|}{}                                                                       & \multicolumn{1}{l|}{A2C}                                       & 0.973                        & {\color[HTML]{F2BA02} 21.026} & -14.099                        & 28.073                        \\
\multicolumn{1}{c|}{}                        & \multicolumn{1}{c|}{}                                                                       & \multicolumn{1}{l|}{DDPG}                                      & 0.973                        & {\color[HTML]{F2BA02} 21.026} & -14.099                        & 28.073                        \\
\multicolumn{1}{c|}{}                        & \multicolumn{1}{c|}{}                                                                       & \multicolumn{1}{l|}{PPO}                                       & —                            & —                             & —                              & —                             \\
\multicolumn{1}{c|}{}                        & \multicolumn{1}{c|}{}                                                                       & \multicolumn{1}{l|}{SAC}                                       & —                            & —                             & —                              & —                             \\
\multicolumn{1}{c|}{}                        & \multicolumn{1}{c|}{\multirow{-5}{*}{FinRL}}                                                & \multicolumn{1}{l|}{TD3}                                       & —                            & —                             & —                              & —                             \\ \cline{2-7} 
\multicolumn{1}{c|}{}                        & \multicolumn{1}{c|}{}                                                                       & \multicolumn{1}{l|}{FinMem (GPT-4o-mini)}                      & -0.09                        & -1.733                        & -14.967                        & 28.26                         \\
\multicolumn{1}{c|}{}                        & \multicolumn{1}{c|}{}                                                                       & \multicolumn{1}{l|}{FinAgent (GPT-4o-mini)}                    & 1.02                         & 17.874                        & -14.519                        & 35.432                        \\
\multicolumn{1}{c|}{\multirow{-16}{*}{MSFT}} & \multicolumn{1}{c|}{\multirow{-3}{*}{LLM}}                                                  & \multicolumn{1}{l|}{\textbf{(Ours)TradingGroup (GPT-4o-mini)}} & {\color[HTML]{00B0F0} 1.655} & 20.273                        & {\color[HTML]{F2BA02} -9.887}  & 21.952                        \\ \hline
\multicolumn{1}{c|}{}                        & \multicolumn{1}{c|}{}                                                                       & \multicolumn{1}{l|}{Buy and Hold}                              & 0.17                         & -12.729                       & -54.402                        & 86.718                        \\
\multicolumn{1}{c|}{}                        & \multicolumn{1}{c|}{}                                                                       & \multicolumn{1}{l|}{SMA Cross}                                 & -0.365                       & -21.327                       & -48.57                         & 59.914                        \\
\multicolumn{1}{c|}{}                        & \multicolumn{1}{c|}{}                                                                       & \multicolumn{1}{l|}{WMA Cross}                                 & 0.245                        & 1.338                         & -38.834                        & 55.764                        \\
\multicolumn{1}{c|}{}                        & \multicolumn{1}{c|}{}                                                                       & \multicolumn{1}{l|}{ATR Band}                                  & {\color[HTML]{F2BA02} 0.933} & {\color[HTML]{F2BA02} 25.169} & {\color[HTML]{F2BA02} -22.906} & \textbf{{\color[HTML]{FF0000} 38.716}} \\
\multicolumn{1}{c|}{}                        & \multicolumn{1}{c|}{}                                                                       & \multicolumn{1}{l|}{Bollinger Bands}                           & -0.795                       & -24.371                       & -40.733                        & {\color[HTML]{00B0F0} 43.904} \\
\multicolumn{1}{c|}{}                        & \multicolumn{1}{c|}{\multirow{-6}{*}{\begin{tabular}[c]{@{}c@{}}Rule\\ Based\end{tabular}}} & \multicolumn{1}{l|}{Turn of The Month}                         & 0.713                        & 21.243                        & \textbf{{\color[HTML]{FF0000} -21.522}} & {\color[HTML]{F2BA02} 49.36}  \\ \cline{2-7} 
\multicolumn{1}{c|}{}                        & \multicolumn{1}{c|}{}                                                                       & \multicolumn{1}{l|}{ARIMA}                                     & —                            & —                             & —                              & —                             \\
\multicolumn{1}{c|}{}                        & \multicolumn{1}{c|}{\multirow{-2}{*}{Predictor}}                                            & \multicolumn{1}{l|}{XGBoost}                                   & {\color[HTML]{00B0F0} 1.52}  & \textbf{{\color[HTML]{FF0000} 84.256}} & -31.901                        & 68.724                        \\ \cline{2-7} 
\multicolumn{1}{c|}{}                        & \multicolumn{1}{c|}{}                                                                       & \multicolumn{1}{l|}{A2C}                                       & —                            & —                             & —                              & —                             \\
\multicolumn{1}{c|}{}                        & \multicolumn{1}{c|}{}                                                                       & \multicolumn{1}{l|}{DDPG}                                      & —                            & —                             & —                              & —                             \\
\multicolumn{1}{c|}{}                        & \multicolumn{1}{c|}{}                                                                       & \multicolumn{1}{l|}{PPO}                                       & —                            & —                             & —                              & —                             \\
\multicolumn{1}{c|}{}                        & \multicolumn{1}{c|}{}                                                                       & \multicolumn{1}{l|}{SAC}                                       & 0.184                        & -13.011                       & -55.542                        & 89.395                        \\
\multicolumn{1}{c|}{}                        & \multicolumn{1}{c|}{\multirow{-5}{*}{FinRL}}                                                & \multicolumn{1}{l|}{TD3}                                       & 0.184                        & -13.011                       & -55.542                        & 89.395                        \\ \cline{2-7} 
\multicolumn{1}{c|}{}                        & \multicolumn{1}{c|}{}                                                                       & \multicolumn{1}{l|}{FinMem (GPT-4o-mini)}                      & -0.58                        & -31.665                       & -47.987                        & 81.186                        \\
\multicolumn{1}{c|}{}                        & \multicolumn{1}{c|}{}                                                                       & \multicolumn{1}{l|}{FinAgent (GPT-4o-mini)}                    & 0.561                        & 3.239                         & -54.625                        & 109.416                       \\
\multicolumn{1}{c|}{\multirow{-16}{*}{COIN}} & \multicolumn{1}{c|}{\multirow{-3}{*}{LLM}}                                                  & \multicolumn{1}{l|}{\textbf{(Ours)TradingGroup (GPT-4o-mini)}} & \textbf{{\color[HTML]{FF0000} 2.051}} & {\color[HTML]{00B0F0} 70.601} & {\color[HTML]{00B0F0} -21.525} & 58.427                        \\ \hline
\end{tabular}
}
\end{table}

\section{Experiments}

\begin{figure*}[!t]
  \centering
  \includegraphics[width=\textwidth]{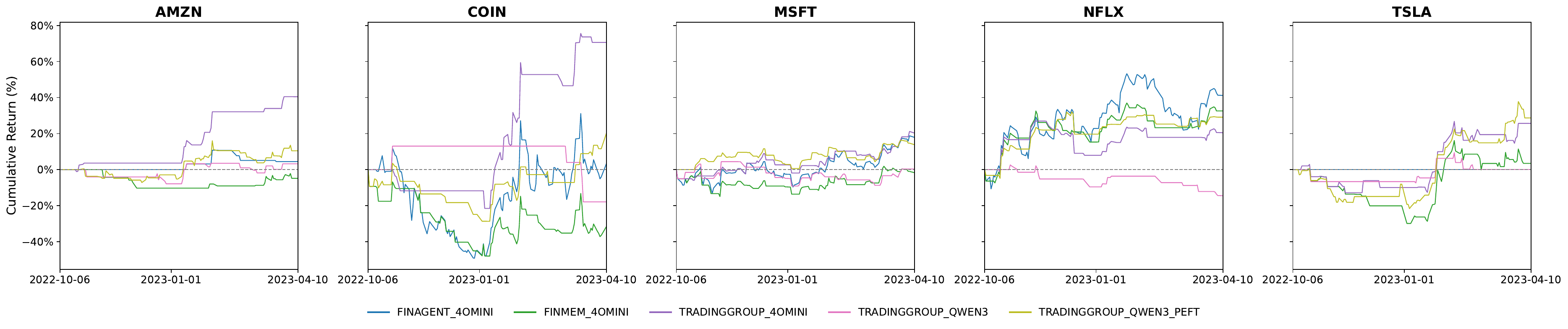}
  \caption{Comparison of cumulative returns across different LLM-based Agents.}
  \label{fig:llm-comparison}
\end{figure*}




\subsection{Datasets}

\noindent\textbf{Training set:} To build instruction data for post-training, we run TradingGroup with DeepSeek-R1 as the base LLM over two nonoverlapping historical windows: 16 June 2020 - 16 August 2021 and 17 August 2021 - 05 October 2022. For each window, we test two modes: risk-management off (fully autonomous decisions) and risk management on (semi-autonomous decisions). During backtesting, the pipeline automatically logs every agent’s inputs, outputs, full CoT, and daily evaluation metrics (see Figure 2). Based on these metrics (mentioned in 3.7), we selected the high-quality decision and forecasting samples from the Trading-Decision Agent and Stock-Forecasting Agent, resulting in 1,080 labeled trajectories that serve as the distillation dataset.

\noindent\textbf{Test set:} We evaluate TradingGroup using the public data provided by the FINSABER backtesting framework~\cite{li2025can} for the period 6 Oct 2022 – 10 Apr 2023, covering five stocks—Amazon.com Inc. (AMZN), Netflix Inc. (NFLX), Tesla Inc. (TSLA), Microsoft Corporation (MSFT), and Coinbase Global Inc. (COIN). The dataset includes closing price, financial news, annual reports and quarterly reports. All five stocks have complete closing price records over 127 trading days. TSLA and MSFT have news articles every day; AMZN has news on only 22 trading days; COIN and NFLX have no news in this window. Except for MSFT (which contains only quarterly filings in this window), the other four stocks include both quarterly and annual reports.

\subsection{Experiments Setup}
To evaluate our TradingGroup and assess the gains from the proposed data-synthesis pipeline, we design two setups
(i)
\textbf{Framework-comparison experiment:} In the first setup, we use GPT-4o-mini~\cite{openai2024gpt4omini} as the unified inference core for all agents and benchmark TradingGroup against four baseline strategies on the same test set, thereby assessing the performance advantage of TradingGroup's multi-agent architecture in trading decisions. To avoid look-ahead bias, all online modules are completely disabled during backtesting.
(ii)
\textbf{Data-synthesis + PEFT experiment:} In the second setup, we uesed TradingGroup’s data pipeline to extract labelled agent decision trajectories from historical periods and perform parameter-efficient fine-tuning (PEFT) of Qwen3-8B using HuggingFace's TRL library. We used the LoRA method combined with int8 quantization, and the trainable parameters were only 0.5301\% of the total parameters. This significantly reduced the computing requirements, enabling our training process to be completed on a single V100 (32G). The AdamW8bit optimiser was used to save memory. Because Qwen3 already possesses strong reasoning capability, we trained for one epoch to avoid overfitting. The entire training process took around six hours. The trained model is called Qwen3-Trader-8B-PEFT. Then we use this trained Qwen3-Trader-8B-PEFT in our TradingGroup to do backtesting on the test set.

\subsection{Baseline and Evaluation Metrics}

To evaluate the effectiveness of our proposed TradingGroup, four typical strategies integrated within the FINSABER framework~\cite{li2025can} are compared.
(1) \textbf{Rule-based strategies:} Buy \& Hold serves as the simplest benchmark, taking a position on the first trading day and holding it until the end; SMA Cross and WMA Cross trade on crossover signals between short-term and long-term moving averages; ATR Band builds a volatility channel from the 14-day ATR, with breakouts of the upper or lower band triggering entry or exit; Bollinger Bands use the 20-day moving average plus or minus two standard deviations to identify overbought or oversold conditions and then reverse; Turn-of-the-Month refers to buying at the close on the last trading day of each month and selling at the close on the first trading day of the next month to capture the month-end effect.
(2) \textbf{Predictor-based strategies:} ARIMA and XGBoost are machine learning strategies that use data-driven model training to forecast future price movements.
(3) \textbf{RL-based strategies:} using the five algorithms of the FinRL~\cite{liu2020finrl} library (A2C, PPO, SAC, TD3, and DDPG) which learn optimal trading policies through interaction with the environment to adapt to market dynamics.
(4) \textbf{LLM-based agents:} FinMem integrates hierarchical memory and role prompts into the agent framework, making the decision-making chain interpretable. FinAgent combines multimodal visual understanding, calls to external tools, and a layered self-reflection module to generate trading actions.

Following FINSABER, 4 metrics are applied for the evaluation: (1) \textbf{Cumulative Return (CR):} Return metric capturing the total profit generated over the backtesting period.
(2) \textbf{Sharpe Ratio (SPR):} Risk-adjusted performance metric that expresses excess return earned per unit of risk.
(3) \textbf{Maximum Drawdown (MDD):} Risk metric measuring the most extreme peak-to-trough loss.
(4) \textbf{Annualised Volatility (AV):} Risk metric describing the volatility of daily returns, scaled annually.

\begin{table}[!t]
\centering
\caption{Performance between Qwen3-Trader-8B-PEFT (fine-tuned with data using TradingGroup's data pipeline) and base Qwen3-8B. Bold indicates better performance.}
\resizebox{\linewidth}{!}{
  \begin{tabular}{clllll}
\hline
\multicolumn{1}{l|}{Symbol}                 & \multicolumn{1}{l|}{Strategy}                            & SPR↑                         & CR(\%) ↑                      & MDD ↑                         & AV(\%) ↓                      \\ \hline
\multicolumn{1}{c|}{}                       & \multicolumn{1}{l|}{TradingGroup (Qwen3-8B)}             & -0.012                       & 0.073                         & \textbf{-8.532} & \textbf{22.963} \\
\multicolumn{1}{c|}{\multirow{-2}{*}{TSLA}} & \multicolumn{1}{l|}{TradingGroup (Qwen3-Trader-8B-PEFT)} & \textbf{1.356} & \textbf{28.666} & -22.17                        & 41.212                        \\ \hline
\multicolumn{1}{c|}{}                       & \multicolumn{1}{l|}{TradingGroup (Qwen3-8B)}             & -2.254                       & -14.49                        & -16.661                       & \textbf{14.732} \\
\multicolumn{1}{c|}{\multirow{-2}{*}{NFLX}} & \multicolumn{1}{l|}{TradingGroup (Qwen3-Trader-8B-PEFT)} & \textbf{1.806} & \textbf{29.11}  & \textbf{-9.341} & 28.905                        \\ \hline
\multicolumn{1}{c|}{}                       & \multicolumn{1}{l|}{TradingGroup (Qwen3-8B)}             & 0.281                        & 3.214                         & \textbf{-7.88}  & \textbf{16.841} \\
\multicolumn{1}{c|}{\multirow{-2}{*}{AMZN}} & \multicolumn{1}{l|}{TradingGroup (Qwen3-Trader-8B-PEFT)} & \textbf{0.849} & \textbf{10.427} & -10.593                       & 22.848                        \\ \hline
\multicolumn{1}{c|}{}                       & \multicolumn{1}{l|}{TradingGroup (Qwen3-8B)}             & 0.035                        & 0.353                         & -12.951                       & 25.295                        \\
\multicolumn{1}{c|}{\multirow{-2}{*}{MSFT}} & \multicolumn{1}{l|}{TradingGroup (Qwen3-Trader-8B-PEFT)} & \textbf{1.311} & \textbf{13.818} & \textbf{-9.858} & \textbf{18.786} \\ \hline
\multicolumn{1}{c|}{}                       & \multicolumn{1}{l|}{TradingGroup (Qwen3-8B)}             & -1.183                       & -17.931                       & \textbf{-27.39} & \textbf{31.608} \\
\multicolumn{1}{c|}{\multirow{-2}{*}{COIN}} & \multicolumn{1}{l|}{TradingGroup (Qwen3-Trader-8B-PEFT)} & \textbf{0.89}  & \textbf{20.078} & -31.102                       & 53.146                        \\ \hline
\end{tabular}
}
\end{table}

\section{Results and Analysis}

\subsection{Performance Comparison}
Table 1 reports four evaluation metrics (CR, SPR, MDD, AV) of TradingGroup (core model: GPT-4o-mini) and all baseline strategies on five stock datasets.  "-" indicates that the strategy never executed any buying action, and thus no trading activity occurred.

On the four datasets of TSLA, AMZN, MSFT and COIN, TradingGroup significantly outperformed all LLM-based baselines (see Table 1 and Figure 3) and achieved the globally optimal overall performance on TSLA, AMZN and COIN. Notably, its cumulative return on AMZN reaches 40.46\%, far exceeding the best baseline's 13.27\%. On the AMZN dataset, TradingGroup has improved its return metrics while still maintaining the lowest maximum drawdown and a relatively favorable annualised volatility.
The combination of ``high return + low risk'' benefits from the introduction of the dynamic risk management module.

Constrained by the risk management threshold, the cumulative return rate of TradingGroup on NFLX is slightly conservative. When the risk management module was turned off (as shown in Table 3), the cumulative return of NFLX increased to 53.24\%, and the Sharpe ratio reached 2.726, ranking first among all methods. However, the risk and return performance of the remaining datasets decreased significantly. Considering the overall stability and the uncertainty of the output of LLMs, the risk management module should be enabled by default.

\subsection{Performance after PEFT}

Table 2 compares the backtesting results between Qwen3-Trader-8B-PEFT and the original Qwen3-8B without using our data-sythesis. The cumulative return curve is also shown in Figure 3. Inference during backtesting was accelerated using vLLM~\cite{kwon2023efficient}.
From the table, we can observe that on all five stocks, the Qwen3-Trader-8B-PEFT outperformed the original Qwen3-8B in both cumulative return (CR) and sharpe ratio (SPR). Among them, the TSLA and NFLX stand out particularly, with CR reaching 28.67\% and 29.11\% respectively, also surpassing the corresponding scores of using GPT-4o-mini (25.66\% and 20.46\%, Table~1).
Further, taking MSFT as an example, Qwen3-Trader-8B-PEFT not only achieves higher CR and SPR, but also reduces both maximum drawdown and annualised volatility, verifying the effective synergy between the TradingGroup’s risk management module and the strategy chain.
Additionally, as the automatic annotation pipeline uses return-related indicators as signals, Qwen3-8B can learn a decision-making thinking chain that is more conducive to improving returns after being fine-tuned with the distillation data from DeepSeek-R1. This also provides an inspiration for the future introduction of a risk-sensitive annotation mechanism in future work.

\subsection{Ablation Studies}

Table 3 shows the contributions of key components of TradingGroup. The Risk-Management module (RM) and Style-Preference Agent \& Current State (PC) are treated as risk-control elements. The Self-Reflection mechanism (SR) here means the LLM-based experiential reflection used by Trading-Decision and Stock-Forecasting, which is considered as the performance optimization element. And Qwen3-Reranker \& Embedding (RE) are considered as the performance optimization elements too. We evaluate four configurations: (i) all four components removed, (ii) only the risk-control elements retained (RM + PC), (iii) all components except RM, and (iv) all components enabled.

On the TSLA dataset, enabling RM + PC cuts cumulative return by 77\% relative to "all removed" but both annualised volatility and maximum drawdown get improvement. By adding RE+SR on the basis of RM+PC, the return and risk indicators can be simultaneously improved in the four datasets of TSLA, AMZN, MSFT and COIN, verifying the important role of retrieval quality and self-reflection in the effectiveness of decision-making. Removing RM (while keeping RE, SR, and PC) can get the largest cumulative return (CR) on NFLX. However, returns fall on the other four datasets, with TSLA dropping to -14.38\%. This shows that completely lifting risk constraints may let the agent deviate excessively in certain markets, upsetting the risk–return balance. Therefore, risk management should be configured flexibly, according to market characteristics and the strategy’s risk tolerance.

\begin{table}[!t]
\centering
\caption{Ablation results across different module combinations (RM: Risk Management, SR: Self Reflection, RE: Qwen3-Reranker \& Embedding, PC: Style-Preference Agent \& Current State). Red means improvement in the metric, while green means reduction.}
\resizebox{\linewidth}{!}{ 
\begin{tabular}{cccccllll}
\hline
\multicolumn{1}{l|}{Symbol}                 & \multicolumn{1}{l}{RM} & \multicolumn{1}{l}{SR} & \multicolumn{1}{l}{RE} & \multicolumn{1}{l|}{PC} & SPR↑                                  & CR(\%) ↑                              & MDD ↑                                  & AV(\%) ↓                              \\ \hline
\multicolumn{1}{c|}{}                       &                        &                        &                        & \multicolumn{1}{l|}{}   & 1.07                                  & 21.904                                & -15.363                                & 42.479                                \\
\multicolumn{1}{c|}{}                       & \checkmark                      &                        &                        & \multicolumn{1}{c|}{\checkmark}  &  0.409\color[HTML]{32CB00}{(-61\%)}   &  5.276\color[HTML]{32CB00}{(-77\%)}   & {-14.144\color[HTML]{FE0000}{(+8\%)}}   & {25.968\color[HTML]{FE0000}{(-39\%)}}  \\
\multicolumn{1}{c|}{}                       &                        & \checkmark                      & \checkmark                      & \multicolumn{1}{c|}{\checkmark}  & {-0.827\color[HTML]{32CB00}{(-177\%)}} & {-14.38\color[HTML]{32CB00}{(-166\%)}} & {-20.379\color[HTML]{32CB00}{(-33\%)}}  & {34.123\color[HTML]{FE0000}{(-20\%)}}  \\
\multicolumn{1}{c|}{\multirow{-4}{*}{TSLA}} & \checkmark                      & \checkmark                      & \checkmark                      & \multicolumn{1}{c|}{\checkmark}  & {1.398\color[HTML]{FE0000}{(+31\%)}}   & {25.662\color[HTML]{FE0000}{(+17\%)}}  & {-15.305\color[HTML]{FE0000}{(+0.4\%)}} & {34.78\color[HTML]{FE0000}{(-18\%)}}   \\ \hline
\multicolumn{1}{c|}{}                       &                        &                        &                        & \multicolumn{1}{l|}{}   & 1.849                                 & 33.139                                & -10.014                                & 32.051                                \\
\multicolumn{1}{c|}{}                       & \checkmark                      &                        &                        & \multicolumn{1}{c|}{\checkmark}  & {0.372\color[HTML]{32CB00}{(-80\%)}}   & {4.792\color[HTML]{32CB00}{(-86\%)}}   & {-9.903\color[HTML]{FE0000}{(+1\%)}}    & {26.383\color[HTML]{FE0000}{(-18\%)}}  \\
\multicolumn{1}{c|}{}                       &                        & \checkmark                      & \checkmark                      & \multicolumn{1}{c|}{\checkmark}  & {2.726\color[HTML]{FE0000}{(+47\%)}}   & {53.244\color[HTML]{FE0000}{(+61\%)}}  & {-11.122\color[HTML]{32CB00}{(-11\%)}}  & {32.102\color[HTML]{32CB00}{(+0.2\%)}} \\
\multicolumn{1}{c|}{\multirow{-4}{*}{NFLX}} & \checkmark                      & \checkmark                      & \checkmark                      & \multicolumn{1}{c|}{\checkmark}  & {1.282\color[HTML]{32CB00}{(-31\%)}}   & {20.458\color[HTML]{32CB00}{(-38\%)}}  & {-14.958\color[HTML]{32CB00}{(-49\%)}}  & {30.131\color[HTML]{FE0000}{(-6\%)}}   \\ \hline
\multicolumn{1}{c|}{}                       &                        &                        &                        & \multicolumn{1}{l|}{}   & 0.601                                 & 7.261                                 & -11.036                                & 22.615                                \\
\multicolumn{1}{c|}{}                       & \checkmark                      &                        &                        & \multicolumn{1}{c|}{\checkmark}  & {0.998\color[HTML]{FE0000}{(+66\%)}}   & {11.391\color[HTML]{FE0000}{(+57\%)}}  & {-14.329\color[HTML]{32CB00}{(-30\%)}}  & {20.789\color[HTML]{FE0000}{(-8\%)}}   \\
\multicolumn{1}{c|}{}                       &                        & \checkmark                      & \checkmark                      & \multicolumn{1}{c|}{\checkmark}  & {0.765\color[HTML]{FE0000}{(+27\%)}}   & {9.411\color[HTML]{FE0000}{(+30\%)}}   & {-17.873\color[HTML]{32CB00}{(-62\%)}}  & {23.072\color[HTML]{32CB00}{(+2\%)}}   \\
\multicolumn{1}{c|}{\multirow{-4}{*}{AMZN}} & \checkmark                      & \checkmark                      & \checkmark                      & \multicolumn{1}{c|}{\checkmark}  & {3.859\color[HTML]{FE0000}{(+542\%)}}  & {40.458\color[HTML]{FE0000}{(+457\%)}} & {-2.118\color[HTML]{FE0000}{(+81\%)}}   & {17.228\color[HTML]{FE0000}{(-24\%)}}  \\ \hline
\multicolumn{1}{c|}{}                       &                        &                        &                        & \multicolumn{1}{l|}{}   & 0.466                                 & 6.362                                 & -12.182                                & 29.107                                \\
\multicolumn{1}{c|}{}                       & \checkmark                      &                        &                        & \multicolumn{1}{c|}{\checkmark}  & {0.733\color[HTML]{FE0000}{(+57\%)}}   & {9.561\color[HTML]{FE0000}{(+50\%)}}   & {-10.105\color[HTML]{FE0000}{(+17\%)}}  & {25.14\color[HTML]{FE0000}{(-14\%)}}   \\
\multicolumn{1}{c|}{}                       &                        & \checkmark                      & \checkmark                      & \multicolumn{1}{c|}{\checkmark}  & {1.423\color[HTML]{FE0000}{(+205\%)}}  & {17.535\color[HTML]{FE0000}{(+176\%)}} & {-12.646\color[HTML]{32CB00}{(-4\%)}}   & {22.368\color[HTML]{FE0000}{(-23\%)}}  \\
\multicolumn{1}{c|}{\multirow{-4}{*}{MSFT}} & \checkmark                      & \checkmark                      & \checkmark                      & \multicolumn{1}{c|}{\checkmark}  & {1.655\color[HTML]{FE0000}{(+255\%)}}  & {20.273\color[HTML]{FE0000}{(+219\%)}} & {-9.887\color[HTML]{FE0000}{(+19\%)}}   & {21.952\color[HTML]{FE0000}{(-25\%)}}  \\ \hline
\multicolumn{1}{c|}{}                       &                        &                        &                        & \multicolumn{1}{l|}{}   & -1.321                                & -22.138                               & -22.138                                & 35.321                                \\
\multicolumn{1}{c|}{}                       & \checkmark                      &                        &                        & \multicolumn{1}{c|}{\checkmark}  & {0.897\color[HTML]{FE0000}{(+168\%)}}  & {21.267\color[HTML]{FE0000}{(+196\%)}} & {-32.02\color[HTML]{32CB00}{(-45\%)}}   & {57.471\color[HTML]{32CB00}{(+63\%)}}  \\
\multicolumn{1}{c|}{}                       &                        & \checkmark                      & \checkmark                      & \multicolumn{1}{c|}{\checkmark}  & {1.457\color[HTML]{FE0000}{(+210\%)}}  & {50.71\color[HTML]{FE0000}{(+329\%)}}  & {-21.065\color[HTML]{FE0000}{(+5\%)}}   & {70.327\color[HTML]{32CB00}{(+99\%)}}  \\
\multicolumn{1}{c|}{\multirow{-4}{*}{COIN}} & \checkmark                      & \checkmark                      & \checkmark                      & \multicolumn{1}{c|}{\checkmark}  & {2.051\color[HTML]{FE0000}{(+255\%)}}  & {70.601\color[HTML]{FE0000}{(+419\%)}} & {-21.525\color[HTML]{FE0000}{(+3\%)}}   & {58.427\color[HTML]{32CB00}{(+65\%)}}  \\ \hline

\end{tabular}
}
\end{table}

\section{Conclusion and Future Work}

We have introduced TradingGroup, an innovative multi-agent system for quantitative trading. By integrating self-reflection, dynamic risk management, stock forecasting, and trade execution, the system outperforms other methods inclduing existing LLM-based agents in FINSABER backtesting. Relying on the end-to-end data synthesis pipeline of TradingGroup, we further distilled and fine-tuned Qwen3-8B to obtain Qwen3-Trader-8B-PEFT. This model not only significantly outperforms the original Qwen3-8B in terms of return metrics, but also exceeds GPT-4o-mini in terms of return metrics on the two datasets, verifying the validity of the synthetic data.

Our future work will focus on three aspects: expanding data and annotation dimensions, designing new evaluation indicators such as risk-control scores, and conducting more fine-grained labeling of agent decisions. We will explore ``three-stage training'' (SFT, GRPO (Group Relative Policy Optimization~\cite{shao2024deepseekmath}) driven by trading returns or price forecasts, Rejection Sampling~\cite{guo2025deepseek}) to further strengthen decision quality. We will also continually improve TradingGroup’s collaboration and safety mechanisms and building an end-to-end multi-agent training platform for quantitative trading.

\bibliographystyle{ACM-Reference-Format}
\bibliography{ACM_Conference_Proceedings_Primary_Article_Template/myrefs}

\end{document}